\PassOptionsToPackage{dvipsnames}{xcolor}
\documentclass{article}
\usepackage{iclr2027_conference,times}

\usepackage{amsmath,amssymb,amsfonts,amsthm,mathtools}

\usepackage[utf8]{inputenc}
\usepackage[T1]{fontenc}
\usepackage[scaled=0.95]{zi4}
\usepackage{CJKutf8}
\usepackage{microtype}
\usepackage{xcolor}
\usepackage{graphicx}
\usepackage{booktabs}
\usepackage{multirow}
\usepackage{array}
\usepackage{nicefrac}
\usepackage{subcaption}
\usepackage{enumitem}
\usepackage{makecell}
\usepackage{colortbl}
\usepackage{pifont}
\usepackage{listings}
\definecolor{mygreen}{RGB}{76, 200, 80}  
\definecolor{mygray}{RGB}{240, 240, 240} 
\definecolor{myred}{RGB}{244, 67, 54}     
\definecolor{bestcell}{HTML}{D8EEE6}
\definecolor{secondcell}{HTML}{F8F1E2}
\newcommand{\vryes}{\textcolor{mygreen}{\ding{51} }}
\newcommand{\vrno}{\textcolor{myred}{\ding{55} }}
\usepackage{listings}

\usepackage{float}
\usepackage{algorithm}
\usepackage{algpseudocode}
\algrenewcommand\Call[2]{\textproc{#1}(#2)}

\usepackage{tikz}
\usetikzlibrary{arrows.meta,positioning,calc,fit,backgrounds,shapes.geometric}
\usepackage{pgfplots}
\usepackage{pgf-pie}
\pgfplotsset{compat=1.17}

\usepackage{hyperref}
\hypersetup{hidelinks}
\usepackage{url}
\usepackage[capitalize,noabbrev]{cleveref}
\makeatletter
\providecommand*{\theHalgorithm}{\arabic{algorithm}}
\providecommand*{\theHALG@line}{\theHalgorithm.\arabic{ALG@line}}
\makeatother
\usepackage[most]{tcolorbox}

\pgfkeysifdefined{/tcb/topfirst}{}{%
  \tcbset{
    topfirst/.style={extras first={top=#1}},
    topmiddle/.style={extras middle={top=#1}},
    toplast/.style={extras last={top=#1}},
  }%
}

\theoremstyle{plain}

\theoremstyle{definition}

\theoremstyle{remark}

\crefname{assumption}{Assumption}{Assumptions}
\Crefname{assumption}{Assumption}{Assumptions}

\newtcolorbox{agentbox}[1][]{
  enhanced,breakable,drop shadow,
  width=\linewidth,
  boxrule=0.8pt,
  colback=SeaGreen!10!CornflowerBlue!10,
  colframe=RoyalPurple!55!Aquamarine!100, 
  left=4pt, right=4pt, top=4pt, bottom=4pt,
  fonttitle=\small\bfseries,
  #1
}

\definecolor{thoughtblue}{HTML}{46468C}
\definecolor{thoughtconblue}{HTML}{F8F8FF}
\definecolor{toolred}{HTML}{B4460A}
\definecolor{toolconred}{HTML}{FFF4EB}
\definecolor{obsgreen}{HTML}{006464}
\definecolor{obscongreen}{HTML}{F0FFFF}

\definecolor{trajectorygreen}{HTML}{41A05F}
\definecolor{trajectorybg}{HTML}{F5FFF5}
\definecolor{stepblue}{HTML}{165696}

\newtcolorbox{trajectorybox}{
    colback=trajectorybg,
    colframe=trajectorygreen,
    boxrule=1.0pt,
    arc=3pt,
    left=10pt,
    right=10pt,
    top=10pt,
    bottom=10pt,
    breakable,
    skin=standard,
    drop fuzzy shadow,
    segmentation hidden,
}

\newtcolorbox{stepheader}{
    colback=white,
    colframe=stepblue,
    colupper=stepblue,
    boxrule=0.8pt,
    arc=2pt,
    left=4pt,
    right=4pt,
    top=2pt,
    bottom=2pt,
    halign=center,
    width=\linewidth,
    before skip=6pt,
    after skip=4pt,
    fontupper=\large\bfseries
}

\newtcolorbox{thoughtblock}{
    colback=thoughtconblue,
    colframe=thoughtblue,
    boxrule=0.8pt,
    arc=2pt,
    left=8pt,
    right=8pt,
    top=8pt,
    bottom=8pt,
    before skip=4pt,
    after skip=6pt,
    width=\linewidth,
    breakable,
    enhanced,
    attach boxed title to top left={yshift=-0.5\baselineskip, xshift=4pt},
    boxed title style={
        colback=thoughtblue,
        colframe=thoughtblue,
        colupper=white,
        boxrule=0pt,
        arc=2pt,
        left=6pt,
        right=6pt,
        top=2pt,
        bottom=2pt,
        fontupper=\small\bfseries
    },
    title=Thought
}

\newtcolorbox{toolblock}{
    colback=toolconred,
    colframe=toolred,
    boxrule=0.8pt,
    arc=2pt,
    left=8pt,
    right=8pt,
    top=8pt,
    bottom=8pt,
    before skip=4pt,
    after skip=6pt,
    width=\linewidth,
    breakable,
    enhanced,
    attach boxed title to top left={yshift=-0.5\baselineskip, xshift=4pt},
    boxed title style={
        colback=toolred,
        colframe=toolred,
        colupper=white,
        boxrule=0pt,
        arc=2pt,
        left=6pt,
        right=6pt,
        top=2pt,
        bottom=2pt,
        fontupper=\small\bfseries
    },
    title=Tool Call
}

\newtcolorbox{obsblock}{
    colback=obscongreen,
    colframe=obsgreen,
    boxrule=0.8pt,
    arc=2pt,
    left=8pt,
    right=8pt,
    top=8pt,
    bottom=8pt,
    before skip=4pt,
    after skip=6pt,
    width=\linewidth,
    breakable,
    enhanced,
    attach boxed title to top left={yshift=-0.5\baselineskip, xshift=4pt},
    boxed title style={
        colback=obsgreen,
        colframe=obsgreen,
        colupper=white,
        boxrule=0pt,
        arc=2pt,
        left=6pt,
        right=6pt,
        top=2pt,
        bottom=2pt,
        fontupper=\small\bfseries
    },
    title=Observation
}

\newcommand{\E}{\mathbb{E}}

\newcommand{\policy}{\pi_\theta}                 
\newcommand{\action}{a}                          
\newcommand{\obs}{o}                             
\newcommand{\query}{q}                           

\newcommand{\goldans}{y^{\star}}                  

\newcommand{\toolname}[1]{\texttt{#1}}
\newcommand{\dataneeded}[1]{}                      

\iclrfinalcopy  

\title{VideoResearchAgent: Grounded Task \\
Synthesis and Sim-to-Real RL for Open-Web \\
Video Research}

\author{%
\parbox[t]{\dimexpr\textwidth-2\tabcolsep\relax}{\centering\normalfont
  \textbf{Yuhang Zhou}$^{1,2}$ \quad
  \textbf{Fei Li}$^{1,2}$ \quad
  \textbf{Yuxi Wu}$^{1,2}$ \quad
  \textbf{Bin Zhu}$^{3}$ \quad
  \textbf{Jingjing Chen}$^{1,2}$\thanks{Corresponding author.} \\
  $^{1}$Institute of Trustworthy Embodied AI, Fudan University \\
  $^{2}$Shanghai Key Laboratory of Multimodal Embodied AI \\
  $^{3}$Singapore Management University \\
  \texttt{ralph.yh.zhou@gmail.com} \quad \texttt{chenjingjing@fudan.edu.cn}
}%
}

\begin{document}

\maketitle
\lhead{Preprint}  

\begin{abstract}

Existing deep research agents are designed primarily for text- and image-based web sources, while video reasoning systems typically assume that relevant videos are provided in advance. We study open-web video research, where an agent must autonomously discover relevant videos, navigate their temporal content, and ground answers in visual evidence. Training such agents at scale is challenging as live video interaction is slow and unreliable, whereas fixed local simulation can induce retrieval-specific shortcuts that fail to transfer to the open web.
We introduce \textbf{VideoResearchAgent}, a scalable training framework to address these challenges. First, we introduce controllable task synthesis pipeline to synthesize multi-hop research tasks from timestamped visual evidence while filtering text-only shortcuts. 
Second, we build a field-aligned local video simulator that preserves deployment-facing search and watch interactions while accelerating video search by 34.5–64.6×. Third, we introduce Retrieval-Domain-Randomized GRPO (RDR-GRPO), which diversifies candidate rankings, distractors, metadata, and result structure during training to reduce overfitting to simulated retrieval.
On
Video-BrowseComp, the VideoResearchAgent trained using Qwen3.5-4B achieves $40.48\%$ accuracy, comparable to
Gemini-3-Flash-Preview, while reducing cumulative API-token consumption by $74.9\%$ relative to the
untrained model. Together, these results establish an accurate and efficient training recipe for
open-web video research. Code, synthetic data, and model weights are available at \url{https://github.com/Ralph-Zhou/VideoResearchAgent}.

\end{abstract}

\section{Introduction}
\label{sec:intro}

Large language models (LLMs) and vision-language models (VLMs) have enabled increasingly capable agents for complex information seeking. Rather than relying solely on parametric knowledge, modern search and deep-research agents can actively formulate queries, inspect external sources, accumulate evidence, and revise their search strategies over multiple steps.
Recent deep research agents have further demonstrated long-horizon planning, multi-hop retrieval, and cross-source synthesis over the open web~\citep{xu2025comprehensive,li2025websailorv2,tongyi2025deepresearch,shao2025drtulu}.

%

However, video remains largely outside this paradigm. As shown in Table~\ref{tab:research_scope}, existing deep research agents predominantly operate over text and static images~\citep{yao2026mmdeepresearch,liu2026pointsseeker,shi2026vidr,zhang2026visualseeker}, whereas video reasoning systems typically assume that the relevant video or video collection is already provided~\citep{gao2025agenticvideo,yang2025longvt,li2026videothinker,lin2026videoseek,zhao2026videoseeker}. This assumption is restrictive for real-world information seeking. Many facts are expressed most naturally through temporal visual evidence, for instance, a transient action in a sports broadcast, a particular step in a tutorial, an object appearing briefly during a product demonstration, or a visual event that is never explicitly described in accompanying text. Answering questions about such evidence requires more than reasoning over a given video. An agent must first find the right video, determine where to look within it, and verify that the observed visual evidence actually supports its answer.


\begin{figure}[t]
    \centering
    \includegraphics[width=0.95\textwidth]{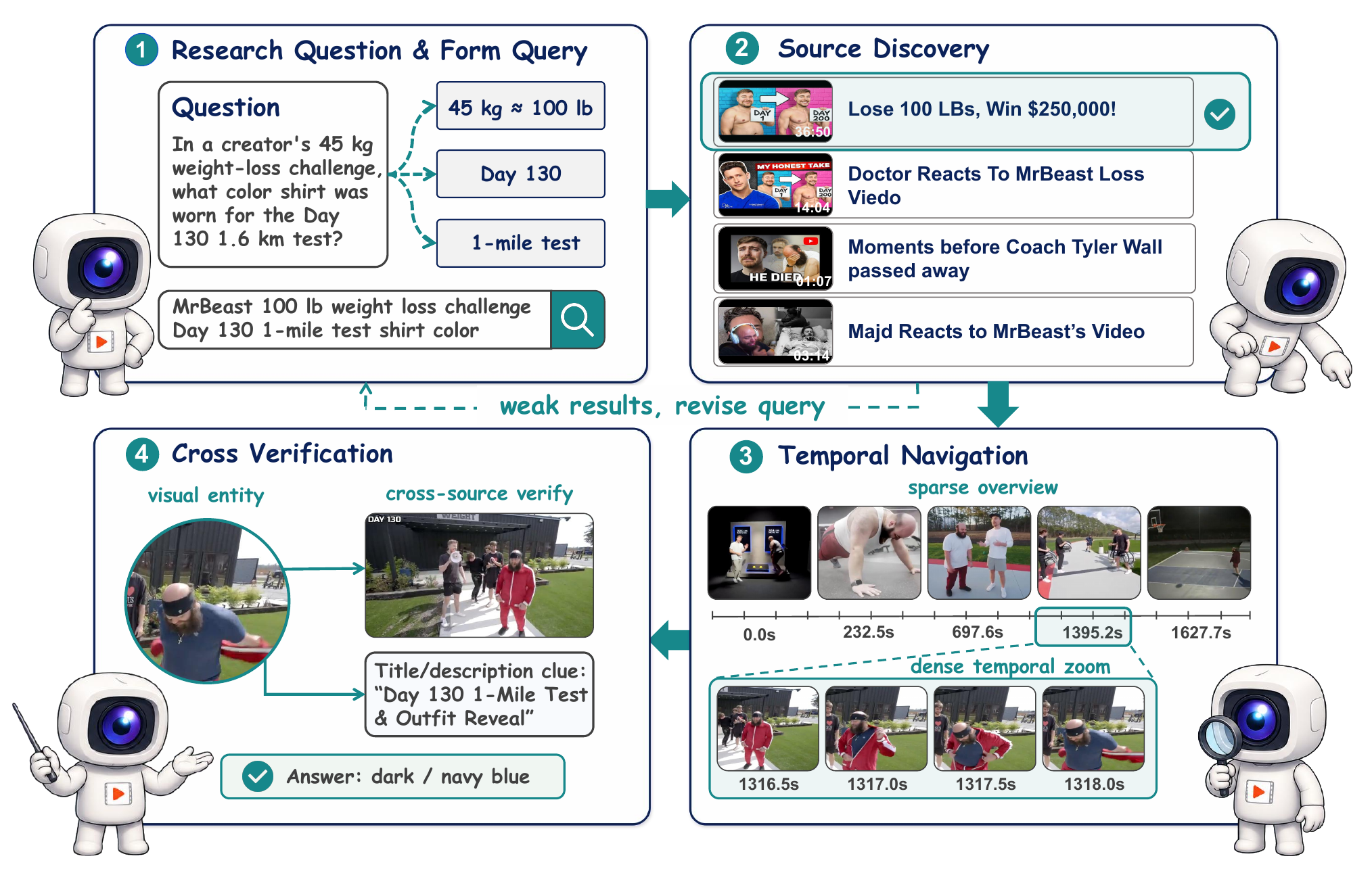}
    \caption{Overview of VideoResearchAgent. A VLM iteratively searches, watches, aggregates, and cross-verifies video evidence.}
    \label{fig:video_research_agent}
    \vspace{-0.3cm}
\end{figure}

%

We study this problem as \textbf{open-web video research}. Given only a natural-language question, an agent must autonomously formulate retrieval queries, discover relevant videos from the open web, navigate their temporal content, and integrate visual evidence across sources before producing an answer.
As illustrated in Figure~\ref{fig:video_research_agent}, 
this setting requires three tightly coupled capabilities: (1) \textit{source discovery} to retrieve useful videos from a large and noisy candidate space, (2) \textit{temporal navigation} to locate informative moments within long and unstructured videos, and (3) \textit{visual verification} to ground the final answer in directly observed evidence. Training such an agent at scale, however, introduces a fundamental challenge between interaction efficiency and deployment fidelity. First, \textbf{existing research-agent training data largely target text or static multimodal browsing, while conventional video datasets provide videos a prior and therefore do not supervise the full process of video discovery and investigation}. Training requires tasks in which the answer genuinely depends on locating and inspecting visual evidence rather than exploiting textual metadata or parametric knowledge. Second, \textbf{performing reinforcement learning directly against live video platforms is prohibitively expensive and unstable}. Video search, retrieval, and decoding incur high latency and are subject to rate limits, network failures, and other external variability. A local video environment can make on-policy training practical, but introduces a third challenge: \textbf{the retrieval distribution inside a simulator is inevitably cleaner and more predictable than that of the live web}. In a fixed local corpus, relevant videos can repeatedly appear at favorable ranks with stable metadata and limited distractors, allowing the policy to exploit simulator-specific retrieval shortcuts rather than learning robust research behavior.

To address these challenges, we introduce \textbf{VideoResearchAgent}, a framework for training agents to conduct open-web video research. First, we develop \textbf{a controllable task-synthesis pipeline} that constructs multi-hop research questions from timestamped visual evidence. The pipeline records explicit evidence provenance and filters tasks that can be solved through closed-book reasoning or text-only search, encouraging genuine video discovery and inspection. Second, we construct \textbf{a field-aligned local simulator} that preserves the deployment-facing video-search and watch interfaces while replacing slow and unreliable live interactions with a high-throughput local backend. The simulator accelerates video search by 34.5--64.6$\times$, making large-scale on-policy training feasible. In addition, we introduce \textbf{Retrieval-Domain Randomization (RDR)} by perturbing candidate rankings, distractors, metadata, and result structure during training. This prevents the policy from exploiting fixed simulator patterns and encourages robust query reformulation, candidate verification, and recovery from retrieval failures. We integrate RDR with Group Relative Policy Optimization (GRPO)~\citep{shao2024deepseekmath}, yielding RDR-GRPO, which trains the policy across diverse retrieval conditions while preserving the deployment-facing action space.



We evaluate VideoResearchAgent on Video-BrowseComp~\citep{liang2025videobrowsecomp}, a benchmark for open-web video search and reasoning. Our VideoResearchAgent framework trained using Qwen3.5-4B substantially improves open-web video research performance. More importantly, controlled post-training improves overall accuracy from 30.00\% to 36.67\% after supervised fine-tuning and further to 40.48\% with RDR-GRPO, reaching performance comparable to Gemini-3-Flash-Preview despite the substantially smaller backbone. Post-training also produces a more efficient research policy: cumulative API-token usage falls by 74.9\% relative to the untrained model, alongside fewer interaction turns and tool calls. Controlled comparisons between fixed-retrieval GRPO and RDR-GRPO further show that retrieval randomization yields stronger improvement under live-web evaluation, supporting its role in reducing simulator-specific overfitting.


\providecommand{\vrpending}{\textcolor{orange!85!black}{\ensuremath{\triangle} }}

\begin{table*}[t]
    \centering
    \caption{Comparison of representative deep-research and video agents. Task synthesis
    denotes a dedicated pipeline for constructing training tasks or tool-use trajectories.
    Under Open Train Data, \vryes indicates a publicly downloadable training set,
    \vrpending an announced but not yet accessible release, and \vrno no such release.
    VideoResearchAgent is the only compared system that combines open-web video search,
    temporal navigation, cross-video evidence, task synthesis, and policy training.}
    \label{tab:research_scope}
    \setlength{\tabcolsep}{4.0pt}
    \renewcommand{\arraystretch}{1.08}
    \resizebox{\textwidth}{!}{%
    \begin{tabular}{@{}lcc ccc ccc@{}}
        \toprule
        \multirow{2}{*}{\textbf{System}} &
        \multicolumn{2}{c}{\textbf{Evidence}} &
        \multicolumn{3}{c}{\textbf{Video Research}} &
        \multicolumn{3}{c}{\textbf{Data \& Learning}} \\
        \cmidrule(lr){2-3}\cmidrule(lr){4-6}\cmidrule(lr){7-9}
        & \makecell{\textbf{Web}\\\textbf{Text}}
        & \textbf{Visual}
        & \makecell{\textbf{Web Video}\\\textbf{Search}}
        & \makecell{\textbf{Temporal}\\\textbf{Navigation}}
        & \makecell{\textbf{Cross-Video}\\\textbf{Evidence}}
        & \makecell{\textbf{Task}\\\textbf{Synthesis}}
        & \makecell{\textbf{Open Train}\\\textbf{Data}}
        & \makecell{\textbf{Policy}\\\textbf{Training}} \\
        \midrule
        \rowcolor{mygray}\multicolumn{9}{c}{\emph{Text-based deep research}} \\
        WebSailor-V2~\citep{li2025websailorv2}
            & \vryes & \vrno & \vrno & \vrno & \vrno & \vryes & \vrno & \vryes \\
        Tongyi DeepResearch~\citep{tongyi2025deepresearch}
            & \vryes & \vrno & \vrno & \vrno & \vrno & \vryes & \vrno & \vryes \\
        DR Tulu~\citep{shao2025drtulu}
            & \vryes & \vrno & \vrno & \vrno & \vrno & \vrno & \vryes & \vryes \\
        MiroThinker~\citep{team2025mirothinker}
            & \vryes & \vrno & \vrno & \vrno & \vrno & \vryes & \vrno & \vryes \\
        OpenSeeker~\citep{du2026openseeker}
            & \vryes & \vrno & \vrno & \vrno & \vrno & \vryes & \vryes & \vryes \\
        OffSeeker~\citep{zhou2026offseeker}
            & \vryes & \vrno & \vrno & \vrno & \vrno & \vryes & \vryes & \vryes \\
        \midrule
        \rowcolor{mygray}\multicolumn{9}{c}{\emph{Image-centric multimodal deep research}} \\
        MM-DeepResearch~\citep{yao2026mmdeepresearch}
            & \vryes & \vryes & \vrno & \vrno & \vrno & \vryes & \vryes & \vryes \\
        POINTS-Seeker~\citep{liu2026pointsseeker}
            & \vryes & \vryes & \vrno & \vrno & \vrno & \vryes & \vrno & \vryes \\
        ViDR~\citep{shi2026vidr}
            & \vryes & \vryes & \vrno & \vrno & \vrno & \vrno & \vrno & \vrno \\
        Visual-Seeker~\citep{zhang2026visualseeker}
            & \vryes & \vryes & \vrno & \vrno & \vrno & \vryes & \vryes & \vryes \\
        Vision-DeepResearch~\citep{huang2026visiondeepresearch}
            & \vryes & \vryes & \vrno & \vrno & \vrno & \vryes & \vrpending & \vryes \\
        Skywork-R1V4~\citep{zhang2025skyworkr1v4}
            & \vryes & \vryes & \vrno & \vrno & \vrno & \vryes & \vrno & \vryes \\
        \midrule
        \rowcolor{mygray}\multicolumn{9}{c}{\emph{Video reasoning}} \\
        AVI~\citep{gao2025agenticvideo}
            & \vrno & \vryes & \vrno & \vryes & \vrno & \vrno & \vrno & \vrno \\
        LongVT~\citep{yang2025longvt}
            & \vrno & \vryes & \vrno & \vryes & \vrno & \vryes & \vryes & \vryes \\
        VideoThinker~\citep{li2026videothinker}
            & \vrno & \vryes & \vrno & \vryes & \vrno & \vryes & \vrno & \vryes \\
        VideoSeek~\citep{lin2026videoseek}
            & \vrno & \vryes & \vrno & \vryes & \vrno & \vrno & \vrno & \vrno \\
        VideoSeeker~\citep{zhao2026videoseeker}
            & \vrno & \vryes & \vrno & \vryes & \vrno & \vryes & \vrpending & \vryes \\
        \midrule
        \rowcolor{mygray}\multicolumn{9}{c}{\emph{Open-web video research}} \\
        Video-Browser~\citep{liang2025videobrowsecomp}
            & \vryes & \vryes & \vryes & \vryes & \vryes & \vrno & \vrno & \vrno \\
        \rowcolor{blue!9}\textbf{VideoResearchAgent (Ours)}
            & \vryes & \vryes & \vryes & \vryes & \vryes & \vryes & \vryes & \vryes \\
        \bottomrule
    \end{tabular}%
    }
    \vspace{-0.3cm}
\end{table*}

Our contributions are summarized as follows:

\begin{itemize}[leftmargin=1.5em,itemsep=2pt,topsep=3pt]
  \item We introduce a scalable training framework for open-web video research, combining evidence-grounded multi-hop task synthesis with a field-aligned local video simulator that enables efficient on-policy training while preserving deployment-facing interactions.
  \item We propose Retrieval-Domain-Randomized GRPO (RDR-GRPO) to address simulator-to-web shift by training policies under diverse rankings, distractors, metadata, and result structures.
  \item Experiments on Video-BrowseComp demonstrate improved accuracy and efficiency.
  Our trained Qwen3.5-4B model achieves accuracy comparable to
  Gemini-3-Flash-Preview, while controlled experiments support the benefit of retrieval randomization for transfer.
\end{itemize}

\section{Related Work}
\label{sec:related}

\textbf{Search and deep research agents.} Search agents extend language models with iterative retrieval, planning, and evidence synthesis. Recent work studies RL-based search, efficient long-horizon interaction, and synthetic trajectory construction in text environments~\citep{jin2025search,li2025websailor,li2025websailorv2,zhou2026offseeker,du2026openseeker,du2026openseekerv2,chu2026redsearcher}. Multimodal extensions further require agents to inspect visual webpages and interleave textual and visual retrieval~\citep{li2025mm,geng2026webwatcher,zhang2026visbrowse,hou2026interlv,wei2026agentic}. Video-BrowseComp~\citep{liang2025video} exposes the next gap: the required evidence may be available only within an open-web video, so locating a page or reading its metadata is insufficient. Our work focuses on training a policy for this search--watch--verify interaction rather than evaluating a fixed research scaffold alone. 

\textbf{Video reasoning.} Long-context and retrieval-augmented video models improve temporal representation and evidence selection~\citep{zhang2024long,chen2025longvila,chen2026eagle,luo2026video,shen2026vgent}. RL-based methods further optimize temporal reasoning or grounding within supplied videos~\citep{feng2025videor1,wang2026videorft,li2025temporalrlt,wang2025timer1}. Deep Video Discovery, VideoSeek, and VideoSeeker introduce active tool use for selecting segments or observations~\citep{zhang2025deepvideo,lin2026videoseek,zhao2026videoseeker}, but operate on a provided video or fixed collection. Open-web video research additionally requires source discovery, timeline navigation, and cross-video verification under noisy retrieval.

\section{Method}

\subsection{Problem Definition}

\label{sec:task}

Open-web video research requires an agent to answer a natural-language question $\query$ by actively seeking evidence from open-web videos. Unlike conventional deep research, which primarily operates over text or static images, open-web video research requires the agent to jointly address \textit{source discovery}, \textit{temporal navigation} and \textit{visual verification} within long and unstructured videos. Specifically, the agent must formulate effective search queries, identify potentially relevant videos, and strategically navigate their temporal content to gather multimodal evidence. Importantly, the agent is not provided with relevant sources, timestamps, or textual surrogates in advance; instead, it must acquire evidence through interaction with the open web, with observed webpage content and video frames serving as its direct evidence for answering the query.

We formulate open-web video search as an interactive sequential decision-making problem. Given a query $\query$, an interaction trajectory is defined as $\tau=(\query,\action_0,\obs_1,\action_1,\obs_2,\ldots,\action_{T-1},\obs_T,\hat y)$, where $\action_t$ denotes the agent's action at step $t$, $\obs_{t+1}$ denotes the resulting textual or visual observation, and $\hat y$ denotes the final answer. The policy $\policy$ and the open-web environment $\mathcal{E}_{\mathrm{web}}$ jointly induce a trajectory distribution $P(\tau\mid\query;\policy,\mathcal{E}_{\mathrm{web}})$. Our objective is to maximize expected final-answer accuracy over the open-web video research task distribution:
\begin{equation}
  \theta^\star = \arg\max_{\theta}
  \E_{(\query,\goldans)\sim\mathcal{D}_{\mathrm{OWVR}}}
  \E_{\tau\sim P(\cdot\mid\query;\policy,\mathcal{E}_{\mathrm{web}})}
  \left[\operatorname{Acc}(\hat y,\goldans)\right],
  \label{eq:objective}
\end{equation}
where $\operatorname{Acc}(\hat y,\goldans)$ measures the semantic correctness of the final answer against the reference answer. 

Despite this formulation, scaling up training for open-web video research remains challenging. We therefore develop a scalable training framework for open-web video research, which we describe in the following sections.

\subsection{Controllable Video-Grounded Task Synthesis}
\label{sec:data}

\begin{figure}[t]
\centering
\includegraphics[width=0.95\textwidth]{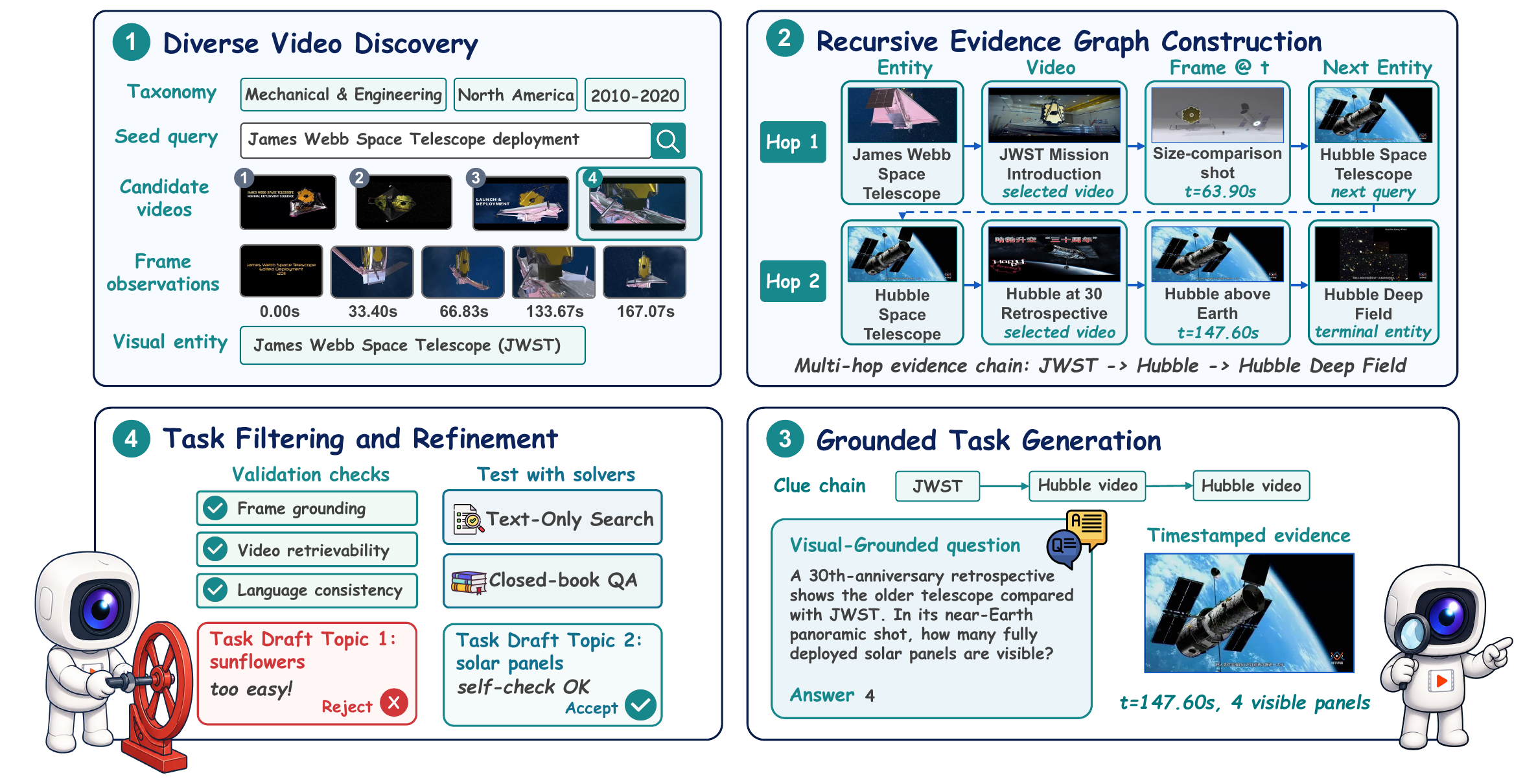}
\caption{Pipeline for synthesizing video-grounded research tasks. Diverse video queries yield visual entities and multi-hop evidence chains, which are turned into questions with concise answers. Task-level
verification reduces text-only shortcuts and increases reliance on video-grounded reasoning.}
\label{fig:task_synthesis_pipeline}
\vspace{-0.3cm}
\end{figure}

Currently, there are no publicly available datasets or established methodologies for constructing open-web video research tasks.
We therefore develop a controllable task synthesis framework that constructs multi-hop evidence chains across open-web videos and converts them into visually grounded research tasks.
This framework enables the effective, large-scale synthesis of open-web video research tasks, supporting model training.
The key idea is to first build a retrievable chain connecting entities, candidate videos, and visual evidence, and then generate questions that require the agent to discover relevant videos and inspect the visual evidence needed to answer the question. To reduce reliance on text-only or parametric shortcuts, we further verify whether the task genuinely requires visual evidence hidden within video frames to solve. As illustrated in \cref{fig:task_synthesis_pipeline}, the framework consists of four stages: (1) diverse video discovery, (2) recursive evidence graph construction, (3) grounded task generation, and (4) task filtering and refinement.

\textbf{Diverse video discovery.}
The first step is to collect initial anchor videos at scale. We employ a two-stage expansion approach: first, an LLM rapidly synthesizes initial seed texts across targeted domains via prompting; second, these seeds are used to locate anchor videos.
To generate seed texts, we sample configurations from a taxonomy spanning domains such as sports, travel, tutorials, product reviews, performances, and documentaries. Each configuration specifies a domain, region and era, which are fed into an LLM to produce a seed query. We then retrieve candidate videos based on availability and duration, using a VLM to describe frames at key timestamps. From these frame observations, a salient visual entity is extracted as the anchor entity to initialize the evidence graph.

\textbf{Recursive evidence graph construction.}
Starting from the anchor entity, we recursively expand the evidence graph, which stores multimodal information required for task generation.
At step $i$, we use the current entity $e_i$ to perform a web search and retrieve candidate videos. We randomly select a video $v_i$ from the search results, sample frames at fixed intervals, and employ a VLM to generate timestamped frame descriptions. A synthesis model then summarizes the video and extracts a salient next entity $e_{i+1}$ from the frame observation $F_i$. This recursive expansion continues until reaching a predefined depth. Consequently, each graph node records the source video, extracted frame descriptions, a video-level summary, and the entity transition. The final graph forms a multi-hop retrievable path across videos, gathering the necessary visual evidence for question synthesis.

\textbf{Grounded task generation.}
Given the multi-hop evidence chain $G=(e_0,v_0,F_0,\ldots,e_k,v_k,F_k)$, we prompt the LLM to generate a question, a concise answer, and the corresponding entity transition path. Questions are crafted such that arriving at the correct answer requires the agent to iteratively search for videos and conduct multi-hop multimodal reasoning. Common question types include counting entities, reading numbers or embedded text, describing spatial relationships, and identifying short events. Each answer is grounded with explicit evidence, including the target entity, supporting video, relevant timestamp, and visual observation.

\textbf{Task filtering and refinement.}
To ensure that the generated tasks genuinely require inspecting visual frame content, we implement a rigorous filtering pipeline. Specifically, we test each task against two baselines: a closed-book LLM and an agent equipped with text-based web search toolkit. Only tasks that cannot be solved by either approach are retained. For surviving tasks, we iteratively rewrite and refine the question to eliminate accidental clues and improve linguistic consistency.
The resulting dataset contains 3,534 tasks. Detailed pipeline and topic distributions are reported in Appendix~\ref{app:data}.

\stepcounter{figure}

\subsection{Supervised Fine-Tuning for Cold-Start} 

For each accepted synthetic task, we provide only the question to the Seed-2.0-Pro~\citep{seed2026seed2} teacher and record its complete trajectory through the live video research interface, including assistant decisions, tool arguments, textual observations, viewed frames, and the final answer. Before training, an independent LLM judge reviews each trajectory for coherent progress, valid and relevant tool calls, avoidance of redundant loops, evidence-grounded answers, and structural correctness. We discard malformed, aimless or unsupported trajectories, yielding 2,301 records; the filtering prompt is provided in Appendix~\ref{app:trajectory_filtering_prompt}. The retained trajectories average 9.32 assistant turns and 8.32 tool calls. 

We perform full-parameter supervised fine-tuning on the filtered trajectories while freezing the vision encoder and multimodal aligner of Qwen3.5-4B. The complete optimization, batching, context, and training configuration is reported in Appendix~\ref{app:sft}.

\subsection{Field-Aligned Local Simulator Construction}
\label{sec:env}

After cold-start, we use RL to further improve the agent's performance on open-web video research tasks. RL rollouts for these tasks trigger frequent video-related tool calls, but live platforms are slow and brittle (suffering from latency, rate limits, HTTP errors, etc.), which destabilizes training. Inspired by ZeroSearch~\citep{sun2025zerosearch}, we build a high-throughput local video backend that preserves the live system's schema and URL structure, while keeping general web search live.

\textbf{Simulator design.}
We simulate two core functionalities: \textbf{video information retrieval} (returning top-$k$ metadata for a search query) and \textbf{video source acquisition} (fetching video files and keyframes for a given path). To this end, we construct a local corpus of 43,751 videos and 1.39 million keyframes based on FineVideo~\citep{Farre2024FineVideo}, assigning each video a stable, deployment-compatible identifier. For retrieval, we merge visual-keyframe and textual-metadata indices, aggregating frame-level similarities to select candidates. For acquisition, the simulator resolves URLs to local files and samples frames across the entire video or a specified timestamp window. Both modules strictly mirror live API schemas, replacing the fragile live backend with a controllable local environment while preserving authentic agent interactions. Additional implementation details are in Appendix~\ref{app:env}.

\label{sec:env_benchmark}

\subsection{Retrieval-Domain-Randomized RL}
\label{sec:rdr_grpo}

We continue to optimize the policy via reinforcement learning (RL) in the environment described in \cref{sec:env}. However, directly training with simulated retrieval can cause policies to overfit to the simulator.
The retrieval distribution inside a simulator is inevitably cleaner and more predictable than that of the live web, which makes it easy for the agent to trust top-ranked results without checking their content or recovering from failed searches.
Such shortcuts transfer poorly to live web search, where relevant videos may be
misranked, mixed with distractors, or missing. This motivated us to consider designing a sim-to-real improvement module to alleviate this issue.



\textbf{Retrieval-Domain Randomization (RDR).}
Prior work on domain randomization shows that policies trained in simulation can better generalize to the real world by \textbf{expanding the distribution of simulated environment to cover the distribution in real-world}~\citep{tobin2017domain}.
Inspired by this, we designed Retrieval-Domain Randomization (RDR) to address the generalization challenge in training with a local simulated retrieval environment.
RDR randomizes the retrieval observations seen during training. Specifically, for a query $q_t$, we first retrieve a candidate pool
$C_t = \mathcal{R}_{\mathrm{sim}}(q_t, K_{\mathrm{pool}})$ and sample a retrieval condition
$z_t \sim p_{\mathrm{RDR}}$ to produce the perturbed observation:
\begin{equation}
  o_t = T_{z_t}(C_t, g, K_t),
  \label{eq:rdr_observation}
\end{equation}
where $g$ contains simulator-side gold candidate identifiers and $K_t$ is the requested
result count. The transformation changes result order, visible metadata, and candidate
availability while preserving valid URLs and watchable video content. The policy observes
only $o_t$, $z_t$, gold identifiers, and internal retrieval scores remain hidden.
This reduces reliance on simulator-specific cues and encourages content verification
and query recovery.
We sample a new $z_t$ for each video-search call, so successive searches can expose the
policy to different retrieval conditions. Implementation details and
mixture parameters are deferred to Appendix~\ref{app:rdr}.

\textbf{RDR-GRPO Training.}
We employ GRPO~\citep{shao2024deepseekmath} for RL training and integrate our RDR module into the local training process. 
To support rollouts in the local environment, we synthesize a separate RL task set and route its video-retrieval calls to the local corpus. This makes the video path controllable, while general web search remains live.
For each trajectory $\tau$, the final reward combines two binary components: an answer correctness score $r_a$, and a format score $r_f$. 
We set $r_a=1$ if an LLM judge finds the final answer semantically consistent with the gold reference, and $r_f=1$ if the final response contains a valid \texttt{<answer>...\allowbreak</answer>} span. The reward is defined as
\begin{equation}
R = r_a + \lambda_f r_f,
\end{equation}
where $\lambda_f = 0.1$.

\section{Experiments}
\label{sec:exp}



\begin{table}[!t]
    \centering
    \caption{Video-BrowseComp results across difficulty levels and video genres. We report
    difficulty-level accuracy, genre accuracy (\%), and overall accuracy (OA). Tool-Free and Search Models results are
    reported from \citet{liang2025videobrowsecomp}; (S) denotes integrated web-search access.
    For the ``VideoResearchAgent with Small-size Models'' group, the highest value in each column
    is shown in bold with green shading, and the second highest is underlined with amber shading.}
    \label{tab:vbc}
    \scriptsize
    \resizebox{0.98\textwidth}{!}{%
    \begin{tabular}{
        @{}
        >{\raggedright\arraybackslash}m{2.6cm}
        >{\centering\arraybackslash}m{0.6cm}
        >{\centering\arraybackslash}m{0.6cm}
        >{\centering\arraybackslash}m{0.6cm}
        >{\centering\arraybackslash}m{0.7cm}
        >{\centering\arraybackslash}m{0.7cm}
        >{\centering\arraybackslash}m{0.7cm}
        >{\centering\arraybackslash}m{0.7cm}
        >{\centering\arraybackslash}m{0.7cm}
        >{\centering\arraybackslash}m{0.7cm}
        >{\centering\arraybackslash}m{0.7cm}
        >{\centering\arraybackslash}m{0.7cm}
        >{\centering\arraybackslash}m{0.7cm}
        @{}
    }
    \toprule
    \multirow{2}{*}{Model} &
    \multicolumn{3}{c}{Difficulty} &
    \multicolumn{8}{c}{Genre Accuracy} &
    \multirow{2}{*}{OA} \\
    \cmidrule(lr){2-4}\cmidrule(lr){5-12}
    & L1 & L2 & L3
    & Doc & Edu & Film & Game
    & Music & Sport & TV & Vid & \\
    \midrule
    \rowcolor{gray!10}\multicolumn{13}{c}{\emph{Tool-Free Models}} \\
    \midrule
    Qwen3-VL-8B-Think
        & 12.00 & 0.00 & 0.00
        & 10.00 & 0.00 & 13.11 & 0.00 & 20.00 & 0.00 & 5.26 & 16.67
        & 7.14 \\
    Qwen3-VL-235B-Ins
        & 22.40 & 0.00 & 0.00
        & 0.00 & 12.00 & 19.67 & 13.64 & 0.00 & 2.22 & 21.05 & 27.78
        & 13.33 \\
    GLM-4.6V
        & 16.80 & 3.23 & 0.00
        & 0.00 & 12.00 & 11.48 & 13.64 & 10.00 & 2.22 & 26.32 & 16.67
        & 10.95 \\
    GPT-4o
        & 28.00 & 3.23 & 0.00
        & 10.00 & 16.00 & 26.23 & 13.64 & 0.00 & 4.44 & 31.58 & 27.78
        & 17.62 \\
    GPT-4o-mini
        & 16.00 & 0.00 & 0.00
        & 0.00 & 12.00 & 18.03 & 9.09 & 0.00 & 0.00 & 15.79 & 5.56
        & 9.52 \\
    GPT-5-mini
        & 26.40 & 0.00 & 0.00
        & 0.00 & 28.00 & 22.95 & 13.64 & 10.00 & 0.00 & 21.05 & 22.22
        & 15.71 \\
    Gemini-2.5-Flash
        & 27.20 & 1.61 & 0.00
        & 20.00 & 20.00 & 24.59 & 4.55 & 0.00 & 2.22 & 36.84 & 22.22
        & 16.67 \\
    Gemini-2.5-Pro
        & 31.20 & 3.23 & 0.00
        & 10.00 & 28.00 & 29.51 & 13.64 & 0.00 & 2.22 & 36.84 & 22.22
        & 19.52 \\
    \midrule
    \rowcolor{gray!10}\multicolumn{13}{c}{\emph{Search Models}} \\
    \midrule
    Gemini-2.5-Flash
        & 32.80 & 4.84 & 0.00
        & 0.00 & 36.00 & 29.51 & 4.55 & 10.00 & 11.11 & 36.84 & 16.67
        & 20.95 \\
    Gemini-2.5-Pro 
        & 37.60 & 4.84 & 0.00
        & 20.00 & 32.00 & 31.15 & 9.09 & 10.00 & 8.89 & 57.89 & 16.67
        & 23.81 \\
    GPT-5.1 
        & 21.60 & 6.45 & 4.35
        & 0.00 & 8.00 & 21.31 & 18.18 & 10.00 & 8.89 & 21.05 & 22.22
        & 15.24 \\
    o4-mini-deep-research
        & 30.40 & 12.90 & 8.70
        & 10.00 & 28.00 & 29.51 & 27.27 & 10.00 & 17.78 & 21.05 & 16.67
        & 22.86 \\
    \midrule
    \rowcolor{gray!10}\multicolumn{13}{c}{\emph{Video-Browser}~\citep{liang2025videobrowsecomp}} \\
    \midrule
    Qwen3
        & 25.60 & 9.68 & 8.70
        & 0.00 & 36.00 & 22.95 & 4.55 & 10.00 & 11.11 & 15.79 & 38.89
        & 19.05 \\
    GPT-5.1
        & 37.60 & 11.29 & 4.35
        & 10.00 & 40.00 & 31.15 & 4.55 & 20.00
        & 20.00 & 26.32 & 44.44
        & 26.19 \\
    \midrule
    \rowcolor{gray!10}\multicolumn{13}{c}{\emph{VideoResearchAgent with Commercial Models}} \\
    \midrule
    Qwen3.5-Plus
        & 53.60 & 29.03 & 26.09
        & 30.00 & 48.00 & 47.54 & 22.73 & 20.00 & 26.67
        & 84.21 & 66.67 & 43.33 \\
    Seed-2.0-Lite
        & 61.60 & 41.94 & 17.39
        & 50.00 & 64.00 & 47.54 & 36.36 & 70.00 & 33.33
        & 78.95 & 66.67 & 50.95 \\
    Seed-2.0-Pro
        & 63.20 & 43.55 & 13.04
        & 50.00 & 56.00 & 54.10 & 40.91 & 60.00 & 40.00
        & 68.42 & 61.11 & 51.90 \\
    GPT-5-mini
        & 40.80 & 20.97 & 8.70
        & 20.00 & 32.00 & 31.15 & 40.91 & 30.00 & 15.56
        & 78.95 & 16.67 & 31.43 \\
    GPT-5.6-Luna
        & 64.00 & 35.48 & 30.43
        & 30.00 & 64.00 & 55.74 & 31.82 & 60.00 & 37.78
        & 78.95 & 61.11 & 51.90 \\
    Gemini-3-Flash-Preview
        & 47.20 & 32.26 & 17.39
        & 20.00 & 64.00 & 36.07 & 27.27 & 20.00 & 33.33
        & 57.89 & 50.00 & 39.52 \\
   
    \midrule
    \rowcolor{gray!10}\multicolumn{13}{c}{\emph{VideoResearchAgent with Small-size Models}} \\
    \midrule
    Qwen3-VL-4B-Think
        & 15.20 & 6.45 & 4.35
        & 0.00 & 28.00 & 14.75 & 0.00 & 20.00 & 8.89 & 0.00 & 11.11
        & 11.43 \\
    Qwen3-VL-8B-Think
        & 26.40 & 4.84 & 0.00
        & 0.00 & 36.00 & 16.39 & 9.09 & 20.00 & 8.89 & 15.79 & 33.33
        & 17.14 \\
    Qwen3.5-9B
        & 43.20 & 17.74 & \cellcolor{secondcell}\underline{17.39}
        & 20.00 & 32.00 & \cellcolor{bestcell}\textbf{50.82} & 13.64 & 20.00 & 8.89 & 57.89 & \cellcolor{secondcell}\underline{44.44}
        & 32.86 \\
    Qwen3.5-27B
        & 45.60 & 19.35 & \cellcolor{secondcell}\underline{17.39}
        & 10.00 & 44.00 & 44.26 & 9.09 & \cellcolor{bestcell}\textbf{60.00} & \cellcolor{bestcell}\textbf{17.78} & 42.11 & \cellcolor{bestcell}\textbf{55.56}
        & 34.76 \\

    Qwen3.5-4B
        & 40.80 & 14.52 & 13.04
        & \cellcolor{bestcell}\textbf{40.00} & 44.00 & 34.43 & \cellcolor{secondcell}\underline{18.18} & 30.00 & \cellcolor{secondcell}\underline{15.56} & 42.11 & 27.78
        & 30.00 \\
    \; + SFT
        & \cellcolor{secondcell}\underline{48.00} & \cellcolor{secondcell}\underline{20.97} & \cellcolor{secondcell}\underline{17.39}
        & 10.00 & \cellcolor{bestcell}\textbf{56.00} & 44.26 & \cellcolor{secondcell}\underline{18.18} & 40.00 & \cellcolor{secondcell}\underline{15.56} & \cellcolor{secondcell}\underline{63.16} & \cellcolor{secondcell}\underline{44.44}
        & \cellcolor{secondcell}\underline{36.67} \\
    \; + RDR-GRPO
        & \cellcolor{bestcell}\textbf{52.80} & \cellcolor{bestcell}\textbf{22.58} & \cellcolor{bestcell}\textbf{21.74}
        & \cellcolor{secondcell}\underline{30.00} & \cellcolor{secondcell}\underline{48.00} & \cellcolor{secondcell}\underline{49.18} & \cellcolor{bestcell}\textbf{27.27} & \cellcolor{secondcell}\underline{50.00} & \cellcolor{bestcell}\textbf{17.78} & \cellcolor{bestcell}\textbf{78.95} & 33.33
        & \cellcolor{bestcell}\textbf{40.48} \\

    \bottomrule
    \end{tabular}
    }
\end{table}
\vspace{-0.3cm}

\subsection{Experimental Setup}
\textbf{Video research agent framework.}
We instantiate the policy as a single VLM that equipped with  \toolname{web\_search}, \toolname{search\_youtube}, \toolname{watch\_video}, and \toolname{visual\_grounding}. 
Among them, \toolname{watch\_video} is used—given a video URL and temporal settings—to output a sequence of frames and potential text information within a specified window, while \toolname{visual\_grounding} is used to further localize elements within an image.
Detailed tool schemas can be found in Appendix~\ref{app:tools}.

\textbf{Benchmark and baselines.}
We evaluate our framework on the Video-BrowseComp benchmark~\citep{liang2025videobrowsecomp}, which
consists of $210$ open-web video search tasks across three difficulty levels. As baselines, we
compare with Tool-Free Models, Search Models, and the Video-Browser system~\citep{liang2025videobrowsecomp}.
Every VideoResearchAgent configuration uses temperature $0$, at most 50 interaction turns, one trajectory per question, and pass@$1$ accuracy. Outputs from our framework are judged by DeepSeek-V4-Flash~\citep{xu2026deepseek}.

\subsection{Main Results}
Table~\ref{tab:vbc} reports the main results on Video-BrowseComp. Key findings are summarized as follows:

\textbf{VideoResearchAgent improves open-web video research accuracy across model backbones.}
For the two backbones that also appear in the Tool-Free group, VideoResearchAgent increases the
overall accuracy of GPT-5-mini from $15.71$ to $31.43$ and Qwen3-VL-8B-Think from $7.14$ to
$17.14$. For GPT-5-mini, the improvement
extends beyond easy questions: L2 and L3 accuracy increase from $0.00$ to $20.97$ and $8.70$,
respectively.

\textbf{SFT and RDR-GRPO progressively improve the Qwen3.5-4B policy.}
The controlled Qwen3.5-4B progression isolates the contribution of our training recipe. SFT raises overall accuracy from $30.00$ to $36.67$, while RDR-GRPO further improves it to $40.48$. 
This simultaneously validates the effectiveness of our data synthesis pipeline, SFT trajectories, and the RDR-GRPO algorithm.
Despite using
a smaller backbone, the final 4B policy surpasses the untrained Qwen3.5-9B and Qwen3.5-27B models by
$7.62$ and $5.72$ points, respectively, and closely matches Gemini-3-Flash-Preview. It also achieves the highest overall accuracy among the evaluated small-size models.

\subsection{Local Video Retrieval Simulator and Live-backend Comparison}
To compare latency, we benchmark the live API backend against our local simulator across 30 sequential video searches and 10 watch-video requests. The live backend caches videos and transcripts so that repeated watch video calls do not require additional network fetches; the fully cached case occurs when both caches hit. We report video search and watch video under cold and fully cached conditions, providing mean successful-call latency, live-call success rate, simulator latency, and the live-to-local latency gap for each.

\begin{table}[t]
  \centering
  \small
  \setlength{\tabcolsep}{5pt}
  \caption{Mean successful-call latency of the live backend and local simulator. SR means success rate.}
  \label{tab:sim_eval}
  \begin{tabular}{@{}
    >{\raggedright\arraybackslash}m{3.6cm}
    >{\centering\arraybackslash}m{2.7cm}
    >{\centering\arraybackslash}m{2.9cm}
    >{\centering\arraybackslash}m{2.2cm}
    @{}}
    \toprule
    \textbf{Operation} & \textbf{Live Backend / SR} & \textbf{Local Simulator / SR} & \textbf{Approx. Gap} \\
    \midrule
    Video search & 12.93 s / 73.3\% & 0.20--0.375 s / 100\% & $34.5$--$64.6\times$ \\
    Watch video (cold) & 41.36 s / 80.0\% & $\sim$7.85 s / 100\% & $5.3\times$ \\
    Watch video (fully cached) & 0.235 s / 100\% & $\sim$0.35 s / 100\% & Comparable \\
    \bottomrule
  \end{tabular}
  \vspace{-0.3cm}
\end{table}

\begin{figure}[t]
    \centering
    \includegraphics[width=0.98\linewidth]{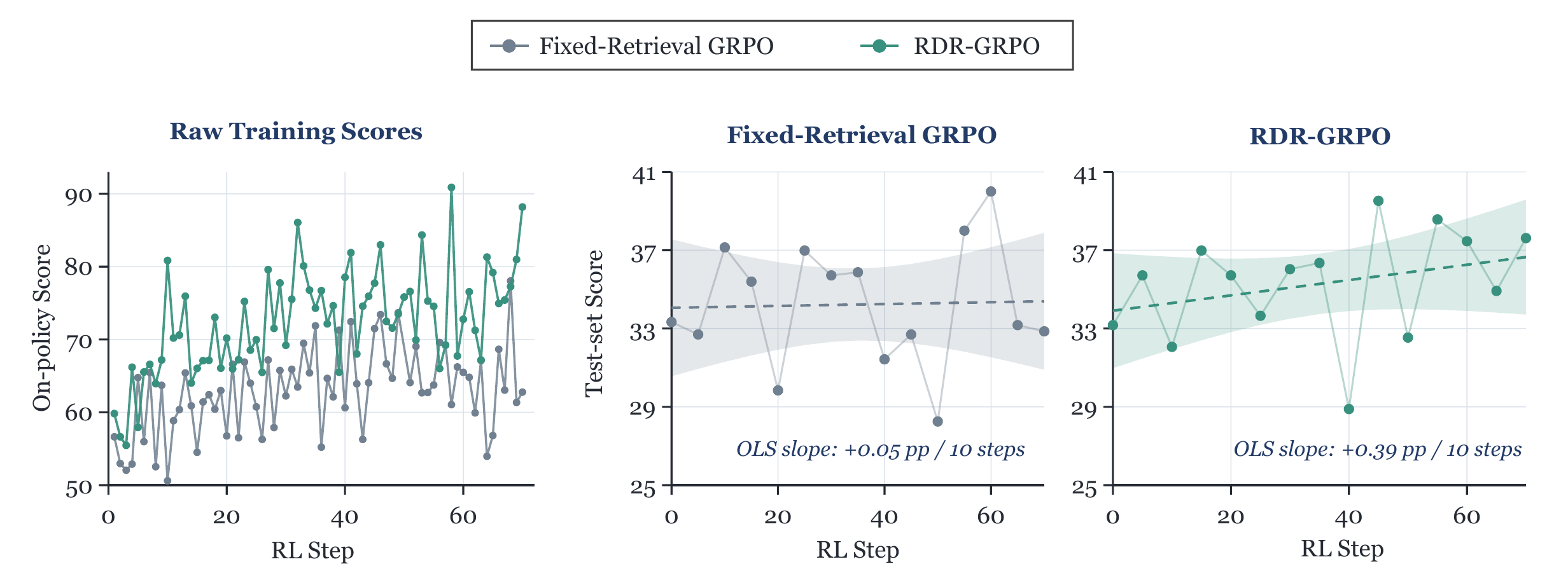}
    \caption{Training and test performance for fixed-retrieval GRPO and RDR-GRPO.
    The left panel shows raw on-policy training scores at each step. The center and right panels show accuracy on test-set data for fixed-retrieval GRPO and RDR-GRPO, with dashed lines and shaded areas indicating linear fits and 95\% confidence intervals.}
    \label{fig:rdr_training_curves}
    \vspace{-0.3cm}
\end{figure}

\begin{figure}[t]
    \centering
    \includegraphics[width=0.98\linewidth]{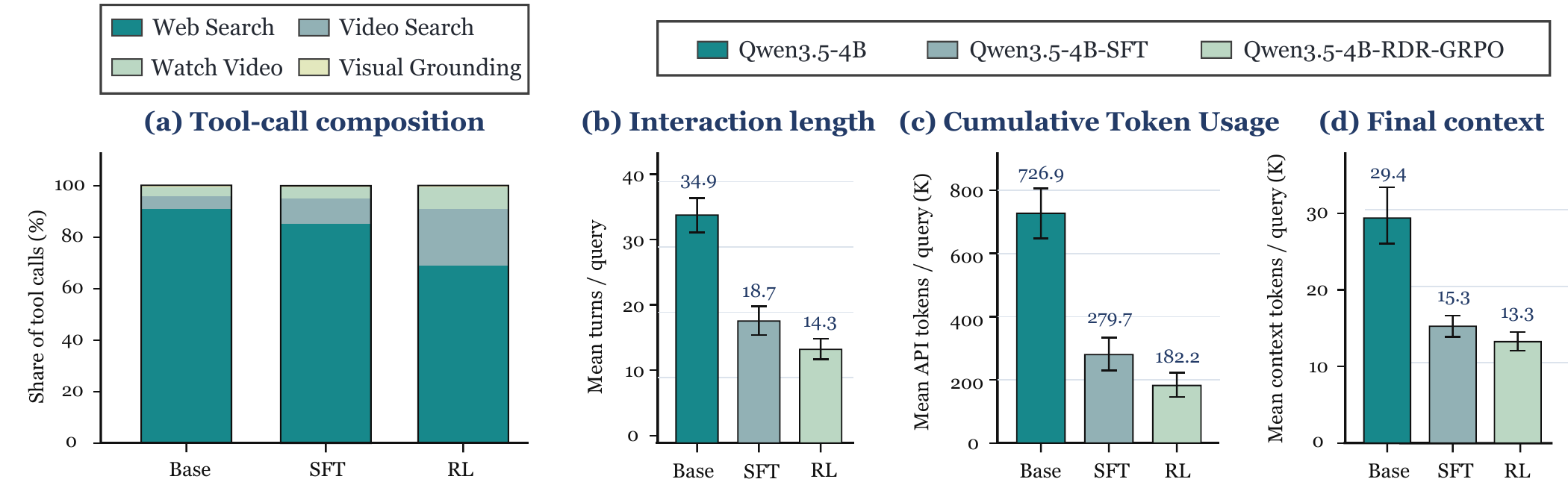}
    \caption{Behavioral statistics for Qwen3.5-4B at the Base, SFT, and SFT+RDR-GRPO stages.
    Panel (a) reports the composition of tool calls and annotates the total calls per query.
    Panels (b)--(d) report mean agent turns, cumulative API token usage, and final serialized context token
    length. Error bars are 95\% task-bootstrap confidence intervals.}
    \label{fig:stagewise_behavior}
    \vspace{-0.2cm}
\end{figure}

\textbf{Results.}
According to \Cref{tab:sim_eval}, the local simulator reduces video search latency from 12.93 seconds to 0.20--0.375 seconds ($34.5$--$64.6\times$ improvement), and cold watch video latency from 41.36 seconds to approximately 7.85 seconds ($5.3\times$ improvement), with 100\% success rates across all evaluated operations (compared to 73.3\% and 80.0\% for live calls). The latency for fully cached watch video calls remains comparable between the two backends. These results show that local simulation significantly reduces end-to-end latency during RL rollouts and improves training efficiency.

\subsection{Retrieval Domain Randomization Ablation}
\label{sec:rdr_ablation}

We compare vanilla GRPO (fixed-retrieval) and RDR-GRPO under identical training settings. 
To save costs and accelerate the process, we randomly sampled 30\% of the tasks from Video-BrowseComp (63 tasks) as the test set.
At matched checkpoints from steps 0--70, both methods are evaluated on the same Video-BrowseComp test split using the live web backend.

\textbf{RDR improves transfer beyond the local retrieval environment.} As shown in \cref{fig:rdr_training_curves}, both methods increase their raw training scores, but vanilla fixed-retrieval GRPO yields a nearly flat test-set trend of $+0.05$ percentage points per 10 steps. RDR-GRPO increases the slope to $+0.39$ under the same training budget, indicating that more of the optimization progress transfers beyond the local retrieval instances.

\subsection{Behavior Analysis}
\label{sec:behavior_analysis}
To understand how our post-training recipe alters policy behavior, we examine matched trajectory checkpoints across all 210 evaluation tasks. We evaluate tool-call composition along with token and interaction consumption, pinpointing the specific mechanisms driving model capability gains.

\textbf{Post-training produces a more video-oriented and resource-efficient research policy.}
As shown in \cref{fig:stagewise_behavior}, progressing from the base Qwen3.5-4B through SFT to RDR-GRPO increases the combined proportion of video search and \toolname{watch\_video} calls from $8.2\%$ to $14.4\%$ and $30.3\%$. Post-training improves task accuracy while cutting cumulative API token consumption per query by $74.9\%$.

\textbf{Control is the primary bottleneck.}
Base models primarily suffer from inefficient interaction control rather than a lack of parametric capacity. The base model frequently fails to converge: $60.0\%$ of its trajectories hit the maximum turn limit, with half degenerating into repetitive tool-calling loops. Scaling model size fails to resolve this bottleneck, as larger models (Qwen3.5-9B/27B/Plus and Gemini-3-Flash) still hit the cap on $24\text{--}36\%$ of tasks. In contrast, post-training reduces the cap hit rate to just $2.9\%$. This trimmed interaction budget is largely redundant: SFT halves total tool calls while preserving nearly the same number of distinct calls ($17.9 \rightarrow 17.2$). Moreover, trajectories lacking evidence after 20 calls rarely find it later ($\le 11\%$), whereas RDR-GRPO maintains active reasoning without abstaining. Finally, upon receiving invalid or low-information search results, the base model repeats the exact same query $35\%$ of the time (Qwen3.5-27B: $18\%$). In contrast, the post-trained policy reformulates the query in $87\%$ of cases, dropping repetition to just $0\text{--}7\%$ across multiple runs. More statistics can be found in Appendix~\ref{app:control}.


\section{Conclusion}


We study open-web video research, requiring agents to discover videos, navigate timelines, and ground answers in visual evidence. To address data, training efficiency, and transfer bottlenecks challenges, VideoResearchAgent introduces controllable task synthesis, a field-aligned local simulator, and RDR-GRPO. On Video-BrowseComp, post-training boosts Qwen3.5-4B accuracy from $30.00\%$ to $40.48\%$ while cutting cumulative API tokens by $74.9\%$. Furthermore, RDR-GRPO exhibits a stronger test-set improvement trend than vanilla fixed-retrieval GRPO, confirming successful policy transfer from simulation to the live web. Overall, these results show that compact VLMs can conduct accurate and efficient open-web video research when training jointly targets temporal supervision, scalable interaction, and generalization.

\bibliography{references}
\bibliographystyle{iclr2027_conference}

\newpage
\appendix
\section{Limitations and Future Work}
\label{sec:limits}

Open-web video research requires long and costly search--watch trajectories, while our RL stage
relies mainly on terminal outcome rewards, providing limited supervision for intermediate research
decisions. Future work could improve sample efficiency through process supervision or finer-grained
credit assignment. In addition, due to resource constraints, we evaluate our training recipe mainly on Qwen3.5-4B. Scaling both model capacity and training data is an
important direction for developing stronger video research agents.

\section{Implementation Details}
\label{app:impl}

VideoResearchAgent uses a dense vision-language policy with native image input and function calling. The
field agent records complete multimodal trajectories in OpenAI-style messages. For RL, these
trajectories initialize a multi-turn verl~\citep{sheng2025hybridflow} training stack with vLLM~\citep{kwon2023efficient} rollout generation and GRPO
updates. Video frames returned by tools are stored as images and reinserted as visual tokens; they
are not serialized into the textual tool response.
The reported RDR-GRPO policy is the final checkpoint produced when training stopped after 85
optimization steps.

\subsection{Teacher Trajectory Filtering}
We employ DeepSeek-V4-Flash~\citep{xu2026deepseek} as a judge to filter the raw teacher trajectories before supervised fine-tuning. The complete judge prompt is shown below:
\label{app:trajectory_filtering_prompt}

\newtcolorbox{judgepromptbox}{
    enhanced,
    breakable,
    colback=gray!5,
    colframe=black,
    boxrule=1.0pt,
    arc=4pt,
    left=8pt, right=8pt,
    bottom=8pt,
    width=\textwidth,
    before skip=10pt,
    after skip=10pt,
    top=18pt,                      
    topfirst=18pt,                 
    topmiddle=-2pt,                 
    toplast=-2pt,                   
    overlay first={
        \fill[black, rounded corners=4pt] 
            ([yshift=0pt] frame.north west) rectangle 
            ([yshift=-20pt] frame.north east);
        \fill[black, sharp corners] 
            ([yshift=-4pt] frame.north west) rectangle 
            ([yshift=-20pt] frame.north east);
        \node[anchor=north west,
              text=white, 
              font=\normalsize\bfseries,
              xshift=8pt,
              yshift=-3pt]
            at (frame.north west) 
            {\textbf{Prompt: Trajectory Quality Judge}};
        \draw[black, line width=0.8pt] 
            ([yshift=-20pt] frame.north west) -- 
            ([yshift=-20pt] frame.north east);
    },
    overlay middle={
        \fill[black, rounded corners=4pt] 
            ([yshift=20pt] frame.north west) rectangle 
            ([yshift=0pt] frame.north east);
        \fill[black, sharp corners] 
            ([yshift=16pt] frame.north west) rectangle 
            ([yshift=0pt] frame.north east);
        \node[anchor=north west,
              text=white, 
              font=\normalsize\bfseries,
              xshift=8pt,
              yshift=17pt]
            at (frame.north west) 
            {\textbf{Prompt: Trajectory Quality Judge}};
    },
    overlay last={
        \fill[black, rounded corners=4pt] 
            ([yshift=20pt] frame.north west) rectangle 
            ([yshift=0pt] frame.north east);
        \fill[black, sharp corners] 
            ([yshift=16pt] frame.north west) rectangle 
            ([yshift=0pt] frame.north east);
        \node[anchor=north west,
              text=white, 
              font=\normalsize\bfseries,
              xshift=8pt,
              yshift=17pt]
            at (frame.north west) 
            {\textbf{Prompt: Trajectory Quality Judge}};
    },
    segmentation hidden=true,
}

\begin{judgepromptbox}
\begin{lstlisting}[
    breaklines=true,
    basicstyle=\ttfamily\scriptsize,
    frame=none,
    xleftmargin=0pt,
    xrightmargin=0pt,
    numbers=none,
    keywordstyle=\bfseries,
    literate={—}{{---}}3 {→}{{$\rightarrow$}}1,
]
You are an expert evaluator for tool-using AI agent trajectories. Your job is to judge whether a single trajectory is **good enough to be used as supervised fine-tuning (SFT) data** for training a multimodal video-research agent.

## Background

- The trajectory you see comes from an agent that answers a video-related
  question by iteratively calling tools such as `search_youtube`, `web_search`,
  `watch_video`, `visual_grounding`.
- The trajectory is the **raw OpenAI-style message log**, with roles
  `system` / `user` / `assistant` / `tool_call` / `tool_response`.
- Image content from tools (e.g. video frames) has been **stripped** and
  replaced by literal `<image>` placeholder tags in the text. **You cannot
  actually see the images.** Each `<image>` tag corresponds to one real frame
  the agent saw in the original run; treat them as opaque visual evidence.
- For each task we provide a `ground_truth` answer that was paired with the
  question when the dataset was synthesised. **This ground truth is NOT
  guaranteed to be correct** — the synthesis pipeline is noisy. Use it only as
  a weak hint; do not penalise a trajectory just because its final answer
  disagrees with `ground_truth`, and do not reward a trajectory just because
  the answer matches.

## What makes a good SFT trajectory

Focus on the **process**, not just the outcome. A keepable trajectory should:

1. **Goal-directed reasoning.** Each step should follow logically from the
   question and prior observations. The agent's `<think>` blocks (if any) and
   tool choices should reflect a coherent search strategy.
2. **Effective tool use.** Tool calls should have well-formed arguments,
   sensible queries, and plausible relevance to the sub-goal. Avoid:
   - obviously malformed JSON or wrong tool names,
   - repetitive identical/near-identical queries that yield nothing new,
   - calling tools whose results the agent never uses.
3. **Productive iteration.** The number of steps should be roughly
   proportional to the question's difficulty. Strongly penalise trajectories
   that loop on the same dead end for many turns, or that hit the iteration
   cap without making concrete progress.
4. **Evidence-grounded final answer.** Whether or not the answer matches the
   (possibly-wrong) ground truth, the answer should be supported by something
   the agent actually observed in tool results. A confident answer pulled out
   of thin air is bad.
5. **Free of structural defects.** No truncated messages, no garbled JSON,
   no role/format inconsistencies, no early crashes.

## What to discard

Mark `keep=false` if any of the following apply:

- The trajectory is dominated by redundant, repetitive, or aimless tool calls.
- The agent runs the loop to the cap (or near it) without producing a real
  answer.
- Tool calls are systematically malformed or hallucinated.
- The reasoning is internally inconsistent (e.g. agent claims to have seen X
  in earlier results but X never appeared).
- The final answer is fabricated with no grounding in tool outputs.
- The trajectory is too short / trivial to teach the model anything (e.g.
  one search → answer, with no real tool use).
- Major formatting / structural issues that would corrupt training.

When in doubt, **be strict**: discarded data is recoverable later, but bad
SFT data poisons the model.

## Output format

Respond with a **single JSON object** and nothing else. The JSON must have
exactly two fields:

{
  "judge_result": "keep" | "discard",
  "reasoning": "<2-5 sentences explaining the judgement, citing concrete signals from the trajectory>"
}

- `judge_result` must be the literal string `"keep"` or `"discard"`.
- `reasoning` must be in the same language as the user's question.
- Do not include any other fields, prefix, suffix, or markdown fences.
\end{lstlisting}
\end{judgepromptbox}
\begin{minipage}{\linewidth}
    \captionsetup{type=figure}
    \caption{LLM judge prompt for filtering teacher trajectories.}
    \label{fig:judge_prompt}
\end{minipage}

\subsection{Supervised Fine-Tuning Configuration}
\label{app:sft}
We use the ms-swift framework~\citep{zhao2024swiftascalablelightweightinfrastructure} for SFT training.
Training uses full-parameter
updates for the language model while keeping the vision encoder and multimodal aligner frozen.
Because multimodal research
trajectories can contain many frames, we set the maximum sequence length to 196,608 tokens and
discard examples that still exceed this limit after multimodal tokenization. The executable settings
are summarized in \cref{tab:sft_hyperparameters}.

\begin{table}[H]
\centering
\caption{Supervised fine-tuning configuration for Qwen3.5-4B.}
\label{tab:sft_hyperparameters}
\small
\setlength{\tabcolsep}{5pt}
\renewcommand{\arraystretch}{1.12}
\begin{tabular}{@{}p{3.9cm}p{8.45cm}@{}}
\toprule
Setting & Value \\
\midrule
Base policy & Qwen3.5-4B \\
Tuning scope & Full-parameter SFT; language model trainable; vision encoder and aligner frozen \\
Training duration & 5 epochs \\
Learning rate & Peak $1\times10^{-5}$; warmup fraction $0.1$; minimum $1\times10^{-6}$ \\
Batching & Micro batch size 1; global batch size 64; length grouping disabled \\
Sequence handling & Maximum length 196,608; padding-free batches; packing disabled; over-length examples deleted \\
Visual input limits & 4,096 image tokens; 128 video tokens; at most 12 sampled video frames \\
Megatron parallelism & Tensor parallel 2; pipeline parallel 1; context parallel 8; sequence parallel enabled \\
Memory optimization & Full uniform activation recomputation (one-layer granularity); complete optimizer CPU offload; precision-aware optimizer enabled \\
Fused kernels & Flash Attention and fused cross-entropy loss \\
Data loading & 8 dataset preprocessing workers and 8 dataloader workers \\
\bottomrule
\end{tabular}
\end{table}

\section{Agent Interface}
\label{app:tools}

The policy operates through four tools and terminates by emitting an answer span. The field and local
implementations share the policy-facing names and semantic outputs.

\begin{table}[H]
\centering
\caption{Policy-facing tools. \toolname{watch\_video} and \toolname{visual\_grounding} return image tokens in addition
to compact textual metadata.}
\label{tab:tools}
\footnotesize
\setlength{\tabcolsep}{4pt}
\begin{tabular}{@{}llp{5.7cm}@{}}
\toprule
Tool & Core arguments & Policy-relevant observation \\
\midrule
\toolname{web\_search}
  & query
  & Ranked web titles, URLs, and snippets for entity discovery. \\
\toolname{search\_youtube}
  & query, top-$k$
  & Ranked video titles, URLs, descriptions, durations, and thumbnails. \\
\toolname{watch\_video}
  & URL, mode, temporal range
  & Transcript context (if available) and either sparse overview frames or dense frames from a selected window. \\
\toolname{visual\_grounding}
  & URL, object prompt, timestamps
  & Localized objects, confidence values, and annotated frames. \\
\bottomrule
\end{tabular}
\end{table}

\subsection{web\_search}

This tool searches the general web for entity discovery and background context. This tool is particularly valuable when the agent lacks prior knowledge of the target video's title or channel—it can gather clues from news articles, Wikipedia, or discussion forums, which are then fed into \texttt{search\_youtube}. We use Serper\footnote{\url{https://serper.dev/}} as the web search engine.

\textbf{Parameters}:
\begin{itemize}
  \item \texttt{query} (string, required): Web search query.
  \item \texttt{max\_results} (integer, optional, default 5): Maximum number of results to return.
\end{itemize}

\textbf{Returns}: A JSON array where each entry contains \texttt{title}, \texttt{url}, and \texttt{snippet}.

\subsection{search\_youtube}

This tool searches YouTube for videos that match a given text query and returns ranked video metadata. It serves as the primary method for video discovery and is usually the initial step in the research process for BrowseComp-style tasks. We primarily use yt-dlp\footnote{\url{https://github.com/yt-dlp/yt-dlp}}, which is lightweight, free, and open-source, to execute YouTube search queries.

\textbf{Parameters}:
\begin{itemize}
  \item \texttt{query} (string, required): YouTube search query.
  \item \texttt{max\_results} (integer, optional, default 10): Maximum number of results to return.
\end{itemize}

\textbf{Returns}: A JSON array where each entry contains \texttt{video\_id}, \texttt{title}, \texttt{url}, \texttt{duration\_seconds}, \texttt{channel}, and \texttt{description}.

\textbf{Implementation}: The tool uses \texttt{yt-dlp} with \texttt{extract\_flat=True} to perform a lightweight YouTube search without downloading video content. Results are ordered by YouTube's relevance ranking. When a cookies file is configured, it is passed to yt-dlp to access age-restricted or region-restricted content.

\subsection{watch\_video}

This tool downloads a video and extracts frames for visual analysis. This is the agent's primary mechanism for acquiring visual evidence—titles and descriptions reveal only surface-level information, while critical answers often reside in fine-grained frame details such as scoreboard numbers, colors, or object counts. Returned frames are injected as visual tokens into the model's multimodal context, enabling reasoning directly from pixel content.

Two operating modes are supported:

\begin{itemize}
  \item \texttt{sparse}: Uniformly samples \texttt{n\_frames} frames across the entire video and returns them together with the full transcript. This mode provides a quick overview and is the default first pass for most tasks due to its lower token cost.
  \item \texttt{dense}: Samples frames at \texttt{fps} within the \texttt{[start\_time, end\_time]} window. This mode is used to zoom into a region of interest after identifying a relevant segment from sparse viewing, enabling fine-grained examination of brief visual details.
\end{itemize}

\textbf{Parameters}:
\begin{itemize}
  \item \texttt{url} (string, required): YouTube video URL or local file path.
  \item \texttt{mode} (string, required): \texttt{"sparse"} or \texttt{"dense"}.
  \item \texttt{start\_time} (number, required for dense mode): Window start time in seconds.
  \item \texttt{end\_time} (number, required for dense mode): Window end time in seconds.
  \item \texttt{n\_frames} (integer, optional, default 16): Number of frames to extract in sparse mode.
  \item \texttt{fps} (number, optional, default 1.0): Sampling rate in frames per second for dense mode.
\end{itemize}

\textbf{Returns}: Text (video duration, extracted frame timestamps, and transcript if available) and images (frames injected as visual tokens).

\textbf{Implementation}: Supports both HTTP(S) URLs and local paths. Videos are cached locally if downloaded, and frames are extracted (GPU-friendly decord preferred). Extracted frames are downscaled (short side $\leq$ 720px), encoded as JPEG base64, and used as visual tokens. If available, transcripts are also included.

\subsection{visual\_grounding}

This tool performs open-vocabulary object detection on video frames. Given a text prompt with period-separated object descriptions, the tool returns bounding boxes with confidence scores and annotated frames. This tool is used when the answer depends on precisely locating, counting, or identifying visual entities that are difficult to discern from sparse frames—for example, counting objects of a specific color or locating a brand logo. Annotated frames are returned as visual tokens, enabling the policy to cross-validate detection results against the original visual context.

\textbf{Parameters}:
\begin{itemize}
  \item \texttt{video\_path} (string, required): Video file path (local path or previously watched YouTube URL).
  \item \texttt{prompt} (string, required): Object descriptions separated by periods.
  \item \texttt{timestamps} (array[number], optional): Specific timestamps in seconds to analyze; if omitted, frames are sampled uniformly from the video.
  \item \texttt{max\_frames} (integer, optional, default 4): Maximum number of frames to analyze.
\end{itemize}

\textbf{Returns}: Text (per-frame detection results: label, confidence score, bounding box coordinates \texttt{[x1,y1,x2,y2]}) and images (frames annotated with bounding boxes and labels, injected as visual tokens).

\textbf{Implementation}: The tool uses Grounding DINO for zero-shot detection~\citep{liu2023grounding}. Results are post-processed to produce image-coordinate bounding boxes. Each detected object is drawn with a colored bounding box and a label showing the class name and confidence score; the annotated frame is encoded as JPEG base64 and returned alongside the text results.

\section{More Task-Synthesis Details}
\label{app:data}

\subsection{Synthesized task distribution}
The topic distribution of the 3,534 synthesized tasks is shown in
\cref{fig:synthesized_task_statistics}.
No single topic accounts for more than 20\% of the corpus. Music \& Performing Arts (19.24\%),
Travel \& Geography (15.85\%), and Sports \& Fitness (13.70\%) are the three largest categories,
jointly accounting for 48.79\% of the tasks. The remaining 51.21\% spans ten categories, indicating
that the synthesized corpus retains broad topical coverage without being dominated by one domain.
\begingroup
\setcounter{figure}{0}
\renewcommand{\thefigure}{A\arabic{figure}}
\renewcommand{\theHfigure}{appendix.\arabic{figure}}
\begin{figure}[t]
\centering
\includegraphics[width=0.95\textwidth]{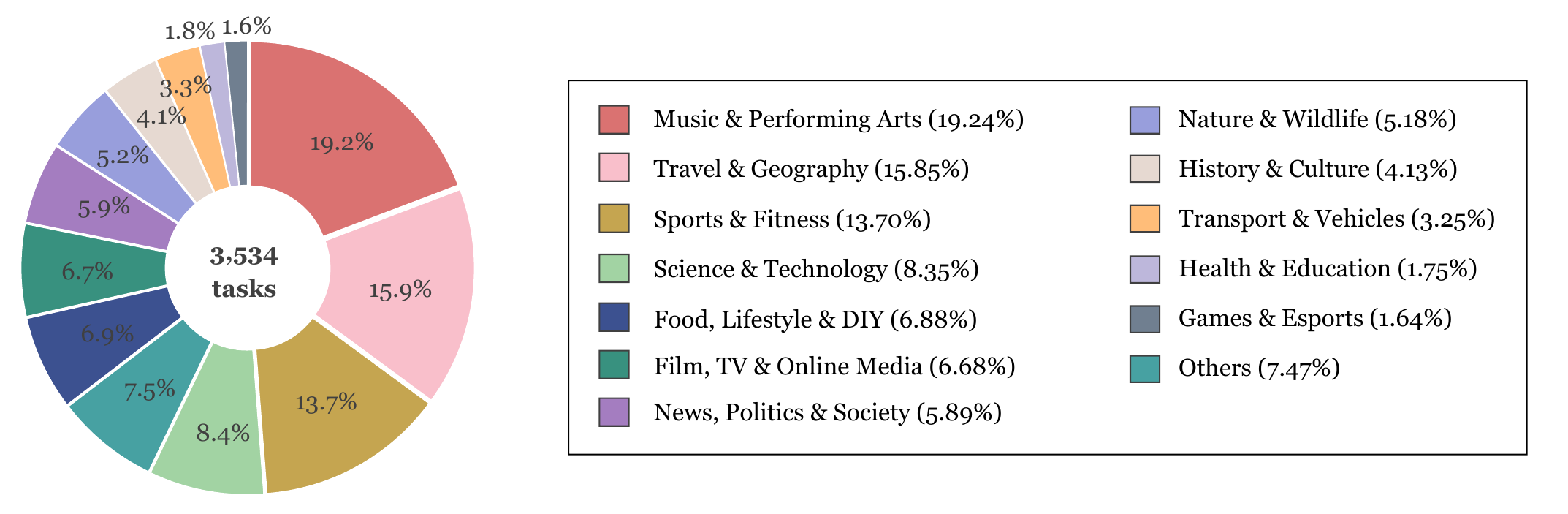}

\caption{Composition of the 3,534 synthesized video research tasks.%
}
\label{fig:synthesized_task_statistics}
\vspace{-0.3cm}
\end{figure}

\endgroup

\subsection{Example synthesized task}
The following accepted task was produced by the online synthesis pipeline from a real evidence graph.

\textbf{Question:}
\begin{quote}
A powerful infrared space telescope, the largest ever launched by humanity, is frequently compared in size to a classic older space telescope that has operated in Earth's orbit for over 30 years. A popular science channel released a 30th-anniversary retrospective of this classic telescope. In the full panoramic shot of the telescope operating in near-Earth space, how many fully visible rectangular solar-panel sections are shown?
\end{quote}

\textbf{Entity chain:}
\begin{itemize}[leftmargin=1.5em,itemsep=2pt,topsep=3pt]
  \item \textit{Seed}: A deployment animation introduces the James Webb Space Telescope (JWST).
  \item \textit{Hop 1}: A NASA mission video compares JWST with the Hubble Space Telescope.
  \item \textit{Hop 2}: A Hubble 30th-anniversary retrospective contains the target operational view.
  \item \textit{Visual evidence}: At $t=147.60$\,s, two rectangular solar-panel sections are visible on each side of Hubble.
\end{itemize}

\textbf{Answer:} \texttt{4}

The task requires the agent to follow a visually induced entity chain from JWST to Hubble, retrieve the corresponding retrospective, and inspect the cited frame. The text-only adversary remained unable to recover the answer after five searches, whereas the frame provides a unique count of four; the task therefore passed the synthesis filters.

\subsection{Data Contamination Analysis}

\paragraph{Audit scope and protocol.}
We audit the 210 Video-BrowseComp evaluation tasks against both the 2,301 filtered trajectories
used for SFT and the complete pre-filtering pool of 3,534 synthesized tasks. The audit combines
normalized exact matching, character $n$-gram TF--IDF, multilingual semantic similarity over both
questions and complete task records, and platform-video identifiers extracted from the full tool
trajectories. We further inspect each
benchmark task's nearest training neighbors under lexical, question-semantic, and record-semantic
similarity, yielding 558 unique pairs, and manually review 88 thresholded high-risk or source-linked
candidates.

\paragraph{Results.}
Across all $210\times2{,}301$ benchmark--SFT comparisons, we find no normalized exact match and no
high-overlap lexical pair: the maximum character TF--IDF similarity is 0.2621, with no pair above
0.70. The maximum question-semantic and record-semantic similarities are 0.8618 and 0.8781,
respectively, with no pair reaching 0.90. The model-assisted nearest-neighbor review labels all 558
pairs as distinct, and manual inspection of the 88 higher-risk candidates finds shared domains or
question templates but no shared target fact. Among the 67 benchmark records with a resolvable
platform video identifier, none of these identifiers is passed to \toolname{watch\_video} in an SFT
trajectory. The single identifier overlap appears only in an unused search result and connects two
different questions. Repeating the checks against the full pre-filtering pool yields the same
conclusion. We also observe 245 answer-only collisions, but these involve common short answers such
as numbers or directions and are not accompanied by question or source overlap.

\paragraph{RL data audit.}
We separately compare all 3,864 unique prompts in the RL training set against the 210 evaluation
tasks, for a total of 811,440 pairs. Normalized exact matching finds no duplicate question, and the
maximum character 3--5-gram TF--IDF similarity is 0.2555, with no pair reaching 0.70. Using the same
multilingual semantic encoder as above, the maximum question similarity is 0.8844 and no pair
reaches 0.90. We manually inspect all 25 pairs with similarity at least 0.86, including the two pairs
above 0.88; they share only broad topics or visual question patterns, not the same video event or
target fact. The RL records also provide explicit gold video URLs. Their 4,569 unique platform-video
identifiers have no intersection with the 70 identifiers resolvable from 67 benchmark records.
Finally, 1,221 answer-only collisions occur across 52 benchmark tasks and 445 RL tasks, but they are
dominated by common colors and numbers and are never accompanied by question or source overlap.

\paragraph{Audit conclusion.}
Taken together, these checks detect no task-level contamination from Video BrowseComp in either
the SFT or RL data. The audit combines exact source matching for 67 of the 210 benchmark tasks with
lexical, semantic, and metadata-level matching across the full evaluation set.

\section{Local Environment Details}
\label{app:env}

\paragraph{Identifier bridge.}
For each local path, the corpus builder computes a deterministic SHA-1 digest, encodes it with
base64url, and projects it to an $11$-character YouTube-style identifier. The resulting URL has the
same surface form as a deployed video URL. A private \texttt{url\_to\_path} map resolves this
identifier to local media for \toolname{watch\_video} and visual grounding. The policy never receives the
local path.

\paragraph{Visual and textual indices.}
The builder extracts configurable uniform or scene-based keyframes and encodes them with
\toolname{clip-vit-large-patch14-336}~\citep{radford2021learning}, producing 768-dimensional CLIP embeddings. Normalized frame
embeddings are stored in a FAISS Flat IP index, while a separate BM25s index covers titles,
descriptions, and subtitles. For a query, dense frame scores are aggregated to the video level, and
the dense and sparse rankings are fused with reciprocal rank fusion,
\begin{equation}
  \operatorname{RRF}(v)=\sum_{m\in\{\mathrm{sparse},\mathrm{dense}\}}
  \frac{1}{k+\operatorname{rank}_{m}(v)}.
\end{equation}
The service returns the top fused candidates under the shared policy-facing schema.

\paragraph{Service and training integration.}
A local FastAPI service exposes the fused retriever and returns each candidate's URL, title, snippet,
duration, and score. A customized tool package connects this service to the training stack.

\paragraph{Hybrid execution.}
During RL, video retrieval, decoding, and visual grounding use the local corpus. General web
retrieval remains live, preserving access to public text context while removing the expensive
video path from the rollout bottleneck.


\section{Retrieval-Domain Randomization Configuration}
\label{app:rdr}

RDR perturbs retrieval observations during training to reduce reliance on the simulator's high recall and stationary rankings. Its goal is to improve transfer to live video retrieval.

\subsection{Difficulty Sampling}

Each \texttt{search\_youtube} call independently samples a difficulty tier from a fixed categorical distribution:

\begin{table}[H]
\centering
\caption{Difficulty tiers and their sampling probabilities.}
\label{tab:rdr_tiers}
\small
\setlength{\tabcolsep}{3pt}
\resizebox{1.0\textwidth}{!}{%
\begin{tabular}{@{}
>{\raggedright\arraybackslash}p{1.2cm}
>{\centering\arraybackslash}p{1.8cm}
>{\raggedright\arraybackslash}p{10cm}@{}}
\toprule
\textbf{Tier} & \textbf{Sampling probability} & \textbf{Purpose} \\
\midrule
easy & 55\% & Preserves low-perturbation cases and a stable training signal. \\
medium & 35\% & Applies moderate rank, distractor, metadata, and structural perturbations. \\
hard & 10\% & Applies stronger perturbations and occasional gold omission or empty results. \\
\bottomrule
\end{tabular}%
}
\end{table}

\subsection{Perturbation Parameters by Difficulty Tier}

\begin{table}[H]
\centering
\caption{RDR perturbation parameters for each difficulty tier.}
\label{tab:rdr_params}
\small
\setlength{\tabcolsep}{3pt}
\resizebox{1.0\textwidth}{!}{%
\begin{tabular}{@{}
>{\raggedright\arraybackslash}p{4.0cm}
>{\raggedright\arraybackslash}p{2.2cm}
>{\raggedright\arraybackslash}p{4.0cm}
>{\raggedright\arraybackslash}p{4.0cm}@{}}
\toprule
\textbf{Perturbation type} & \textbf{easy} & \textbf{medium} & \textbf{hard} \\
\midrule
Rank intervention & gold 100\% keep at top & gold 50\% keep, 35\% sink to middle, 13\% sink to bottom, 2\% drop & gold 25\% keep, 30\% sink to middle, 33\% sink to bottom, 12\% drop \\
Hard negative injection & 0 & 2--4 & 4--8 \\
Title jitter probability & 0\% & 40\% & 80\% \\
Snippet jitter probability & 0\% & 40\% & 80\% \\
Empty result probability & 0\% & 1\% & 3\% \\
Display count variation & none & $-$1 to +2 & $-$2 to +3 \\
\bottomrule
\end{tabular}%
}
\end{table}

\subsection{Four Types of Perturbation Operations}

\paragraph{Rank intervention.}
The gold video's rank position is reassigned according to one of four actions: \texttt{keep} retains the gold video at the top; \texttt{sink\_mid} moves it to the middle (approximately the $K/2$-th position); \texttt{sink\_bottom} moves it to the bottom (approximately the $(K-1)$-th position); and \texttt{drop} removes the gold video from the display window entirely. The probability distribution over these actions for each difficulty tier is shown in \cref{tab:rdr_params}.

\paragraph{Hard negative injection.}
Non-gold videos are sampled from the tail of the same query's candidate pool and promoted into the display window. These candidates remain watchable and provide query-relevant distractors.

\paragraph{Metadata corruption.}
Titles and snippets are stochastically rewritten: title transformations include adding clickbait prefixes, uppercasing, truncating to 40\%--80\% of original length, and appending exclamation marks or emojis; snippet transformations include truncating to 30\%--70\% of original length, lowercasing, and hard truncation to 15--40 characters. The jitter probability for each tier is shown in \cref{tab:rdr_params}.

\paragraph{Structural noise.}
With a certain probability, the tool returns an empty result set or varies the number of displayed candidates. These perturbations expose the policy to search failures and variable result counts. Their probabilities and ranges are shown in \cref{tab:rdr_params}.

\subsection{Sampling Granularity and Safety Constraints}

A new difficulty tier is sampled for each \texttt{search\_youtube} call. The random seed combines a base seed, the rollout instance identifier, and the call index. Repeated searches within one rollout can therefore receive different perturbations.

RDR changes candidate membership, order, count, and displayed metadata. It never synthesizes or rewrites URLs. Promoted negatives come from the local candidate pool and remain resolvable by \texttt{watch\_video}.




\section{Interaction Control Analysis}
\label{app:control}

This section reports more statistics for \cref{sec:behavior_analysis}.

\paragraph{Definitions.}
A trajectory is \emph{capped} if it uses all 50 turns, after which the harness demands a final answer.
A \emph{degenerate loop} is a run of at least five consecutive identical tool calls (same tool and
normalized arguments). An \emph{exact repeat} is a call identical to an earlier call of the same
trajectory, and \emph{distinct calls} are the unique calls of a trajectory. A trajectory
\emph{acquires answer-bearing evidence} when it watches an answer-bearing video (the gold source
video, or the final evidence video of a correct trajectory in another evaluation run) or retrieves
text that contains the reference answer; this gives a lower bound. After a search that returns empty
results, we classify the next call as an identical repeat (the same query up to case, punctuation, and
word order), a minor edit (token-level Jaccard similarity $\geq 0.7$), or a reformulation (any other
search query).

\begin{table}[htbp]
\caption{Interaction statistics of the three Qwen3.5-4B stages on the 210 tasks. Abstention: the final
answer states that the answer cannot be determined.}
\label{tab:control_stages}
\centering\small
\begin{tabular}{lccc}
\toprule
 & Base & SFT & RDR-GRPO \\
\midrule
Accuracy (\%)                                       & 30.0    & 36.7  & 40.5 \\
Trajectories hitting the 50-turn cap (\%)           & 60.0    & 15.7  & 2.9 \\
Trajectories with a degenerate loop (\%)            & 50.5    & 0.0   & 0.0 \\
Tool calls per task                                 & 35.8    & 17.7  & 13.3 \\
Distinct tool calls per task                        & 17.9    & 17.2  & 13.1 \\
Exact-repeat tool calls (\%)                        & 50.1    & 2.8   & 2.0 \\
Abstentions / empty answers (\#)                    & 12 / 18 & 0 / 1 & 0 / 0 \\
API tokens per query (K)                            & 726.9   & 279.7 & 182.2 \\
\midrule
\multicolumn{4}{l}{\emph{Still running without answer-bearing evidence after 20 tool calls}} \\
Trajectories (\#)                                   & 90      & 59    & 36 \\
Later acquire evidence (\%)                         & 1.1     & 8.5   & 11.1 \\
Their calls after the 20th (\% of all tool calls)   & 35.6    & 30.2  & 15.4 \\
\bottomrule
\end{tabular}
\end{table}

\paragraph{Qwen3.5-4B stages.}
60.0\% of base trajectories hit the cap and 50.5\% contain a degenerate loop, so half of all base tool
calls exactly repeat an earlier call (\cref{tab:control_stages}). SFT halves the tool calls per task
($35.8\rightarrow17.7$) while keeping the number of distinct calls nearly unchanged
($17.9\rightarrow17.2$). Calls issued late rarely pay off: among trajectories that still lack
answer-bearing evidence after 20 calls, only 1.1--11.1\% acquire it later, yet the base model issues
35.6\% of its calls in them after the 20th call. RDR-GRPO leaves fewer such trajectories
($90\rightarrow36$) and has no abstentions or empty answers.

\begin{table}[htbp]
\caption{Interaction control of the tested model policies. 50-turn cap: trajectories that
hit the cap, and those among them that contain a degenerate loop. After an empty search result: the next tool
call after a search that returns no results; in the remaining 2--8\% of cases the policy switches to a
video tool or answers.}
\label{tab:control_agents}
\centering\small
\setlength{\tabcolsep}{4pt}
\begin{tabular}{lccccccc}
\toprule
 & & \multicolumn{2}{c}{50-turn cap (\%)} & & \multicolumn{3}{c}{After empty search (\%)} \\
\cmidrule(lr){3-4}\cmidrule(lr){6-8}
Policy & Acc.\ (\%) & All & w/ loop & Calls/task & Reform. & Minor edit & Repeat \\
\midrule
\multicolumn{8}{l}{\emph{Untrained Qwen3.5, increasing scale}} \\
Qwen3.5-4B                    & 30.0 & 60.0 & 49.5 & 35.8 & 38 & 24 & 35 \\
Qwen3.5-9B                    & 32.9 & 35.7 & 24.3 & 26.0 & 54 & 9  & 34 \\
Qwen3.5-27B                   & 34.8 & 23.6 & 12.1 & 21.6 & 68 & 10 & 18 \\
Qwen3.5-Plus                  & 43.3 & 26.2 & 0.0  & 25.0 & 92 & 2  & 0 \\
\midrule
\multicolumn{8}{l}{\emph{Post-trained Qwen3.5-4B (ours)}} \\
+ SFT                         & 36.7 & 15.7 & 0.0  & 17.7 & 92 & 4  & 0 \\
+ RDR-GRPO                    & 40.5 & 2.9  & 0.0  & 13.3 & 87 & 8  & 0 \\
\midrule
\multicolumn{8}{l}{\emph{Other agents}} \\
Gemini-3-Flash                & 39.5 & 27.6 & 5.2  & 26.3 & 28 & 45 & 25 \\
GPT-5-mini                    & 31.4 & 5.2  & 0.0  & 16.1 & 86 & 7  & 0 \\
Seed-2.0-Pro                  & 51.9 & 1.9  & 0.0  & 11.1 & 88 & 7  & 0 \\
Seed-2.0-Lite                 & 51.0 & 1.4  & 0.0  & 11.8 & 93 & 2  & 1 \\
GPT-5.6-Luna                  & 51.9 & 0.5  & 0.0  & 22.0 & 82 & 13 & 1 \\
\bottomrule
\end{tabular}
\end{table}

\paragraph{Across policies.}
Scaling reduces non-convergence but does not remove it (\cref{tab:control_agents}). The untrained
Qwen3.5-9B, 27B, and Plus hit the cap on 35.7\%, 23.6\%, and 26.2\% of tasks, and Gemini-3-Flash on
27.6\%, whereas post-training brings the 4B model to 2.9\%, close to Seed-2.0 and GPT-5.6-Luna
(0.5--1.9\%). After an empty search result, the base model repeats the identical query 35\% of the time
(Qwen3.5-9B: 34\%, Qwen3.5-27B: 18\%), whereas SFT and RDR-GRPO reformulate 92\% and 87\% of failed
queries and repeat fewer than 1\%.

\section{Trajectory Example}
\begin{figure}[t]
\renewcommand{\thefigure}{A\arabic{figure}}
    \centering
    \includegraphics[width=\linewidth]{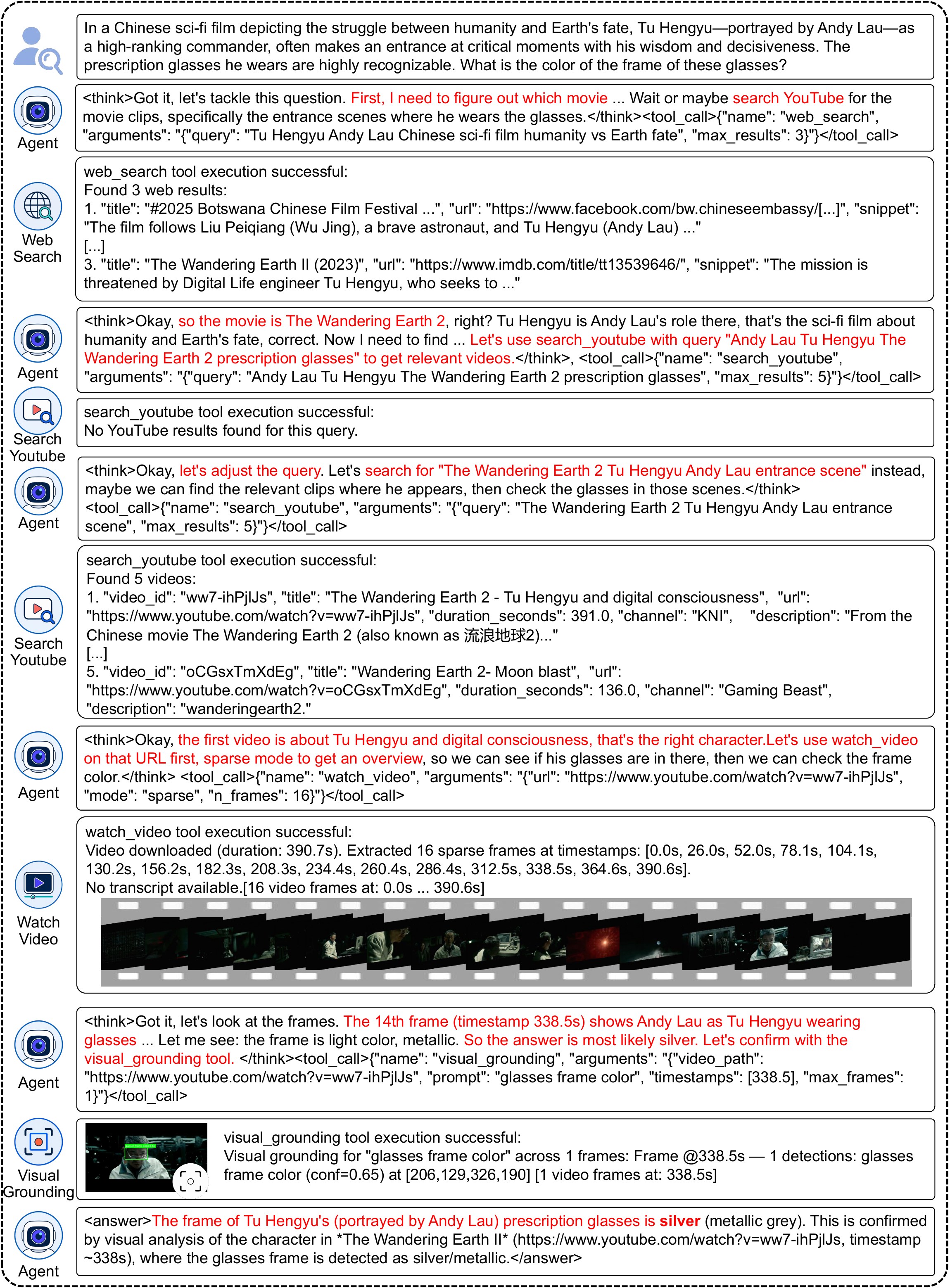}
    \caption{\textbf{An illustrative reasoning trajectory of \textsc{VideoResearchAgent}}. The trajectory details the agent's internal thought process, tool invocations, and observations, ultimately culminating in the correct final answer. Text highlighted in \textcolor{red}{red} denotes pivotal steps along the agent's reasoning path, illustrating how it progressively gathers evidence to reach the final answer.}
    \label{fig:example_0924}
\end{figure}

\end{document}